\documentclass[dvipsnames,sigconf,anonymous=false,review=false,preprint,authorsversion]{acmart}

\renewcommand\footnotetextcopyrightpermission[1]{}
\setcopyright{none}

\AtBeginDocument{%
  }

\usepackage{microtype}
\usepackage{cleveref}
\usepackage{algorithm}
\usepackage[noend]{algpseudocode}
\usepackage{mathtools}

\newcommand{\blfootnote}[1]{%
  \begingroup
  \renewcommand{\thefootnote}{}%
  \footnotetext{#1}%
  \endgroup
}

\definecolor{red}{RGB}{175, 42, 20}
\definecolor{green}{RGB}{109, 158, 75}
\definecolor{blue}{RGB}{76, 115, 187}
\definecolor{purple}{RGB}{90, 64, 154}

\begin{document}

\title{A Path to Universal Neural Cellular Automata}


\author{Gabriel B\'ena}
\email{g.bena21@imperial.ac.uk}
\orcid{https://gabrielbena.github.io/}
\affiliation{%
    \institution{Imperial College}
    \city{London}
    \country{UK}
}

\author{Maxence Faldor}
\email{m.faldor22@imperial.ac.uk}
\orcid{https://maxencefaldor.github.io/}
\affiliation{%
    \institution{Imperial College London}
    \city{London}
    \country{United Kingdom}
}

\author{Dan Goodman}
\email{d.goodman@imperial.ac.uk}
\orcid{https://neural-reckoning.org/dan_goodman.html}
\affiliation{%
    \institution{Imperial College London}
    \city{London}
    \country{United Kingdom}
}

\author{Antoine Cully}
\email{a.cully@imperial.ac.uk}
\orcid{https://www.imperial.ac.uk/personal-robotics/people/previous-post-doctoral-researchers/antoine/}
\affiliation{%
    \institution{Imperial College London}
    \city{London}
    \country{United Kingdom}
}

\renewcommand{\shortauthors}{Béna et al.}

\begin{abstract}
    Cellular automata have long been celebrated for their ability to generate complex behaviours from simple, local rules, with well-known discrete models like Conway’s Game of Life proven capable of universal computation. Recent advancements have extended cellular automata into continuous domains, raising the question of whether these systems retain the capacity for universal computation. In parallel, neural cellular automata have emerged as a powerful paradigm where rules are learned via gradient descent rather than manually designed. This work explores the potential of neural cellular automata to develop a continuous Universal Cellular Automaton through training by gradient descent. We introduce a cellular automaton model, objective functions and training strategies to guide neural cellular automata toward universal computation in a continuous setting. Our experiments demonstrate the successful training of fundamental computational primitives --- such as matrix multiplication and transposition --- culminating in the emulation of a neural network solving the MNIST digit classification task directly within the cellular automata state. These results represent a foundational step toward realizing analog general-purpose computers, with implications for understanding universal computation in continuous dynamics and advancing the automated discovery of complex cellular automata behaviours via machine learning.\blfootnote{Web version available at \url{https://gabrielbena.github.io/blog/2025/bena2025unca/}}
\end{abstract}





\begin{teaserfigure}
    \centering
    \includegraphics[width=0.99\linewidth,keepaspectratio]{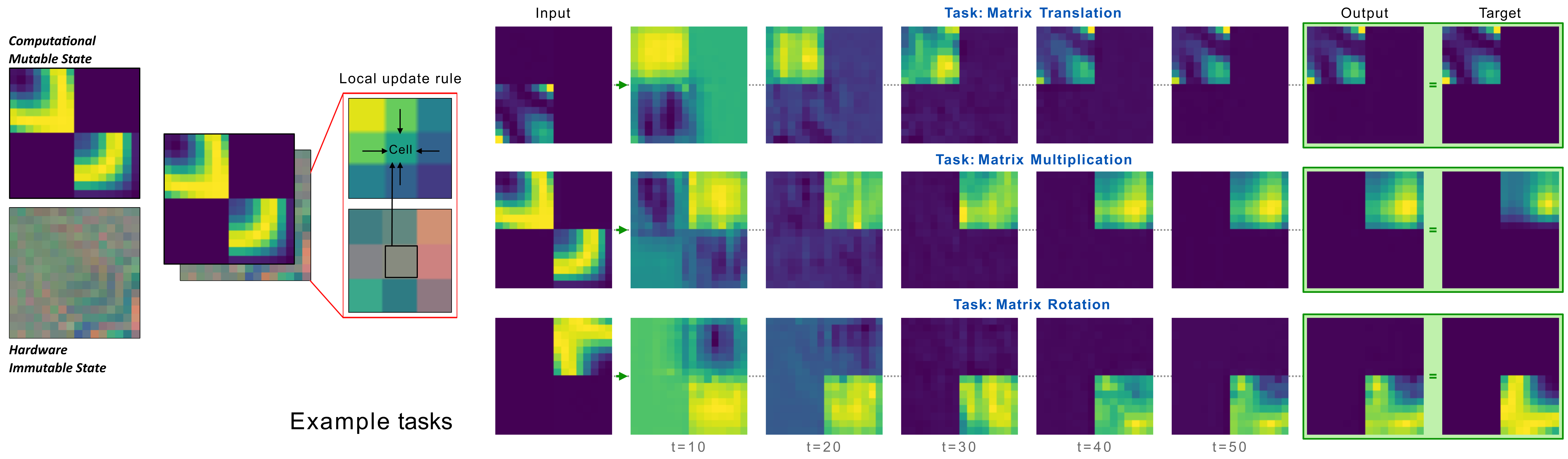}
    \caption{Our framework performs matrix operations (translation, multiplication, rotation) through continuous cellular interactions evolving over time (from $t=0$ to $t=60$). The inset reveals the core mechanism: cells communicating only with neighbors, and interacting with a local fixed (but learned) heterogeneous substrate, are able to collectively solve mathematical tasks without explicit algorithms.}
    \label{fig:teaser}
\end{teaserfigure}


\maketitle

\pagestyle{plain}

\section*{Contributions}

Gabriel Béna and Maxence Faldor contributed equally to this work. Maxence Faldor initiated the project, conceived its central ideas, including the concept of state and immutable state for hardware representation. He also developed the initial software library and experimental design. Gabriel Béna provided crucial feedback on these foundational ideas, developed the tasks setup, led the development of the compiler and the Graph Neural Network (GNN) components, conducted the extensive experiments, and was the primary author of the project website. Dan Goodman and Antoine Cully supervised the project, offering guidance, insightful discussions, and contributing to brainstorming sessions.

\newpage
\section{Introduction}
Cellular Automata (CA) represent a fascinating class of computational models that have captivated researchers across disciplines for their ability to produce complex behaviours from simple rules~\citep{computational_beauty}.
At their core, CA typically consist of a grid of cells, each in one of a finite number of states, which evolve over discrete time steps based on a fixed deterministic rule.
This rule, applied uniformly to all cells, governs the state transition of each cell based solely on its current state and those of its neighbors.
Despite their simplicity, CA have become a cornerstone in studying emergence and complexity~\citep{wolfram_cellular_1984}.
Mathematical proofs have established that well-known CA, such as Conway’s Game of Life, Elementary Cellular Automata, and Wireworld, are capable of universal computation, underscoring their remarkable power and expressiveness~\citep{cook,life_turing_2016}.
Beyond these formal demonstrations, researchers have constructed fully functional Turing machines within these CA, albeit through arduous efforts requiring meticulous design and substantial time investment.

In recent years, the development of continuous CA has enabled to bridge the gap between the discrete nature of traditional models like Conway’s Game of Life and the analog characteristics of the real world.
Notable examples include Lenia~\citep{lenia} and SmoothLife~\citep{smoothlife}, which extend classic CA to simulate lifelike patterns with continuous dynamics.
However, a key open question persists: are these models capable of universal computation? While the answer is likely affirmative, given their expressive potential, proving this remains elusive.
The lack of discrete states and well-defined transitions makes it harder to encode symbolic information reliably --- slight perturbations can lead to significant divergence over time, undermining the stability required for computation.
Moreover, in contrast to the sharp boundaries and interactions of traditional CA, continuous models often exhibit smooth, fuzzy dynamics that make it challenging to design modular components like wires, gates, or memory elements with predictable behavior.
What was already a laborious task in discrete CA becomes more difficult, if not practically impossible, in the continuous domain, highlighting a fundamental challenge in building efficient analog computers.

In parallel with these developments, Neural Cellular Automata (NCA) have emerged as a compelling paradigm that blends the local, decentralized dynamics of classical CA with the representational capacity and trainability of neural networks~\citep{nca}.
Unlike traditional CA, where the update rule is explicitly handcrafted, NCA leverage differentiable architectures wherein the rule is parameterized by a neural network and optimized end-to-end via gradient descent. This makes it possible to \emph{learn} complex behaviours and dynamics from data, bypassing the need for manual rule design.
Recent work has demonstrated that NCA can be trained to perform a wide range of tasks: from self-organizing into complex morphologies~\citep{nca}, to solving algorithmic reasoning tasks such as instances of the 1D-ARC challenge~\citep{cax}, exhibiting emergent collective behaviours~\citep{self_classifying}, and growing artificial neural networks~\citep{hypernca}. These results highlight the versatility of NCA as a model of computation and pattern formation.
Given the Turing completeness of classical CA, NCA offer an exciting new lens through which to explore the space of rules --- not by manually engineering them, but by searching for them through optimization. In essence, NCA turn rule discovery into a machine learning problem.
This shift is significant: the traditionally arduous task of hand-crafting rule sets that give rise to desired behaviours is now offloaded to the learning algorithm itself. Backpropagation through time, combined with the differentiable nature of NCA, allows for flexible, automated tuning of highly non-trivial dynamics.

In this work, we explore the potential of the Neural Cellular Automata paradigm to pioneer the development of a continuous Universal Cellular Automata~\citep{universal_ca}, with the ambitious goal of inducing a universal Turing machine~\citep{universal_turing_machine} to emerge within a continuous CA through training via gradient descent.
The development of Universal Neural Cellular Automata has implications beyond academic curiosity or creating interesting simulations. It touches on fundamental questions about the potential for continuous dynamic systems to exhibit universal computation, or the possibility to create a universal analog computer.
This paper establishes the foundational framework and presents promising initial steps toward realizing this grand vision.
First, we propose a novel framework that disentangles the concepts of ``hardware'' and ``state'' within NCA. In this abstraction, CA rules serve as the ``physics'' dictating state transitions across space and time, akin to the fundamental laws governing computation. The CA state, in turn, acts as the dynamic physical substrate --- comparable to electrical charges in a computer or neurochemical patterns in a brain—while the ``hardware'' represents an immutable scaffold, fixed in its spatial configuration throughout a simulation. This hardware can be leveraged by the physics (i.e., the CA rules) to guide computation but remains unalterable during runtime, providing a stable backbone for emergent behaviours.
Second, we introduce preliminary objective functions and training setups designed to steer NCA toward universal computation in a continuous domain.
Third, we conduct experiments demonstrating the training of essential computational building blocks --- such as matrix multiplication, dot-product and transposition --- within the NCA framework.
Finally, we showcase the practical utility of these building blocks by emulating a neural network directly within the mutable CA state, successfully solving the MNIST digit classification task.
These results mark a critical first step, illustrating how NCA can harness gradient descent to sculpt continuous CA into powerful, general-purpose computational systems.

\section{Related Work}
Discrete CA have long been a cornerstone for studying universal computation due to their ability to generate complex behaviours from simple, local rules. A number of discrete CA, such as Conway’s Game of Life~\citep{cgol,life_turing_2016}, Rule 110 of Elementary Cellular Automata~\citep{wolfram,cook}, Langton’s Ant~\citep{langton}, and Wireworld~\citep{wireworld} have all been proven Turing-complete. These works rely on discrete states and labor-intensive, hand-crafted designs. To mitigate this, evolutionary algorithms have been employed to automate the discovery of CA rules or patterns with specific properties~\citep{mitchell,sapin}, reducing human effort though still targeting discrete systems and predefined goals rather than general-purpose computation.

The shift to continuous CA aims to connect discrete models with real-world analog systems, prompting inquiries into their computational potential. Models like Lenia~\citep{lenia} and SmoothLife~\citep{smoothlife} introduce smooth state transitions, yielding lifelike emergent patterns, yet their capacity for universal computation remains unproven. Recent efforts have applied evolutionary search to Lenia~\citep{reinke_intrinsically_2020,leniabreeder} to automatically discover and optimize patterns, though these pursuits prioritize specific behaviours over general-purpose computation. Similarly, gradient descent has been used on Lenia patterns and rules to discover and optimize patterns~\citep{hamon_discovering_2024}, though again not targeting general-purpose computation.

NCA mark a paradigm shift by replacing hand-crafted rules with neural networks trained via gradient descent. NCA have been successfully applied to specific tasks such as morphogenesis~\citep{nca}, classification~\citep{self_classifying}, and solving difficult problems like the 1D-ARC challenge~\citep{cax}. Most relevant to our work, HyperNCA~\citep{hypernca} uses NCA to grow artificial neural networks, suggesting broader computational versatility. These advances highlight NCA’s strength in automating rule discovery, offloading the burden of manual design to machine learning. Nevertheless, prior NCA research predominantly focuses on pattern formation or specific tasks. It is to be noted that the matrix copy and multiplication tasks were first implemented with NCA by Peter Whidden \citep{comp_nca}, and that our approach builds upon this idea that NCA can be used in a continuous computational setting. Very recently, \citep{difflogic_nca} have also demonstrated that NCA rules can be implemented using (differentiable) logic gates, confirming the possibility to run self-organizing systems on standard digital harware. Finally, in parallel of our work \citep{engram_nca} was developped using a similar idea: cells augmented with "private memory tapes" were shown to exhibit stable, multi-task capabilities.

Our research intersects with analog computing and particularly neuromorphic approaches that bridge biology and silicon. Analog computation utilizes continuous physical systems, and leverages physical phenomena such as wave propagation, diffusion, and material properties to represent and transform information continuously, avoiding the quantization overhead of digital systems \citep{Ulmann+2022}. NCAs draw inspiration from biological neural networks, which primarily employ local computations for energy efficiency \citep{small_world, modular_brain, Bullmore2012TheEO}. This principle of locality is fundamental to both systems, and demonstrates that sophisticated computation can emerge from simple, localized rules without requiring global connectivity.
The brain's co-location of computation and memory offers a solution to the von Neumann bottleneck that increasingly limits conventional computing systems as model complexity grows  \citep{bottlneck}. Neuromorphic systems implement this biological principle through distributed processing elements with local memory, often using mixed-signal circuits that approximate neural dynamics while maintaining energy efficiency \citep{neuromorphic_mead, neuromorphic_review, schuman2017surveyneuromorphiccomputingneural}.
While deep learning has flourished through hardware-software co-design optimized for parallel matrix operations, this specialization has simultaneously restricted algorithmic innovation to operations aligned with current hardware capabilities. By exploring how systems with local interactions like NCAs can implement universal computation, we can develop more versatile computing architectures that maintain the locality constraints of biological systems while leveraging silicon's speed advantages. This approach may yield computing systems that better balance computational power with the remarkable efficiency and adaptability characteristic of biological intelligence.

\section{Methods}
We leverage the CAX~\citep{cax} library for high-performance neural cellular automata implementions and run our experiments on a single L40 GPU.

\subsection{General Setup}
\label{subsec:setup}
The goal of our framework is to demonstrate the ability of neural cellular automata to act as a general computational substrate. To do so, we tackle a variety of tasks, directly in the NCA state (see \ref{subsec:tasks}). In this work, we introduce a novel architecture design that enhances such computational capabilities. Our approach partitions the NCA state space into two distinct components.

\begin{itemize}
    \item The mutable state: this serves as the main workspace (where the tasks inputs are transformed into their outputs). This state is the only one changing through time during a single experiment / task roll-out. This is the computational substrate, meaning that tasks (such as matrix operations) are directly embedded in this space, and transformations on inputs have to happen in this state. The update dynamics of the mutable state are governed by the NCA rules, described in \cref{subsec:nca}.
    \item The immutable state: this functions as a specialized hardware configuration which is spatially heterogeneous. This hardware can itself be monolithic (the same shape as the entire grid) or modular (created from  different specialized components for each task specific instance). This is learned across training but fixed in any duration of a single experiment / task instance. Details are specified in \cref{subsec:hw}
\end{itemize}

Overall, this framework enables a two-level optimization strategy:
At the global level, we train a general-purpose NCA rule (perceive and update functions) to support diverse computational operations. At the task-specific level, we optimize individual hardware configurations.
The system achieves task-specific computation by adapting its dynamics using the local available hardware (reminiscent of placing the correct components on a motherboard).
From an efficiency perspective, this architecture also provides significant practical advantages: once the general NCA rule is trained, adapting the system to new tasks requires only the optimization of hardware configurations, a process that is substantially less computationally intensive than training the full NCA rule from scratch. Full training details are described in \cref{subsec:training}.

\begin{figure}[h!]
    \centering
    \includegraphics[width=0.99\linewidth]{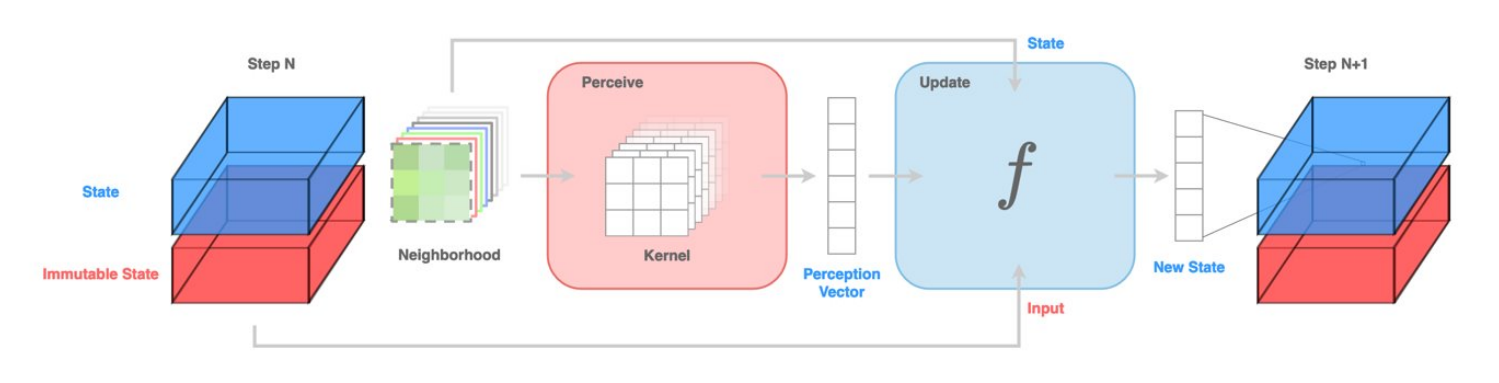}
    \caption{Schematic of our architecture, showing the distinction between mutable (computational) and immutable (hardware) states. }
    \label{fig:architecture}
    \Description{}
\end{figure}

\subsection{Neural Cellular Automata}
\label{subsec:nca}
We develop computational models that solve tasks directly within their mutable state, with dynamics governed by local cell interactions. These interactions are parametrized by a neural network serving as the cellular automaton's rules. The two key components of this neural network are its perception function and update function:

\textbf{Perceive Function: }
The perceive function gathers information about each cell's neighborhood through learnable convolution filters applied to the immediate vicinity within the mutable state. This perception module transforms input state channels into a higher-dimensional perception vector, capturing relevant local spatial patterns. The kernel size, padding, and activation functions are configurable hyperparameters.

\textbf{Update Function: }
The update function utilizes the perception vector and local hardware vector to update each cell's state. Our architecture employs an attention-based update module that calculates state update $\Delta S$ by conditioning perceived information $P$ on an external input vector $I$, which encodes task-specific information or global context (in our implementation, representing local cell hardware).


\begin{enumerate}
    \item  Each cell receives a perception vector $P$ (local spatial patterns) and a hardware vector $I$ (its immutable state).
    \item The hardware vector $I$ activates different "computational modes" through an attention mechanism: $\alpha = \text{softmax}((I \cdot W_{\text{embed}}) / T)$, where $W_{\text{embed}}$ is a learned embedding matrix and $T$ is a temperature parameter controlling the sharpness of the activation.
    \item The perception vector $P$ is simultaneously processed through $N$ parallel pathways (implemented as MLPs), producing potential update vectors $V_h$ for each pathway.
    \item The final state update is computed as a weighted mixture of these pathways: $\Delta S = \sum_{h=1}^{N} \alpha_h V_h$, with the cell's state updated residually: $S_{t+1} = S_t + \Delta S$.
\end{enumerate}

This design allows the NCA to adapt its behavior dynamically based on the local hardware—cells in input regions might activate different computational pathways than those in output regions or computational zones. The result is a flexible computational substrate where the same underlying rule can perform diverse operations depending on the hardware context.

\subsection{Hardware (Immutable State)}
\label{subsec:hw}
As briefly explained in \cref{subsec:setup}, a core innovation in our approach lies in separating \textit{mutable} and \textit{immutable} parts of the computational state. This distinction separates the update model's role (which needs maximal generality and expressiveness) from task-specific hardware configurations that can be diverse and fine-tuned. We explore two different approaches for designing these specialized hardware configurations.

\subsubsection{Monolithic hardware}
\begin{figure}
    \centering
    \includegraphics[width=0.99\linewidth]{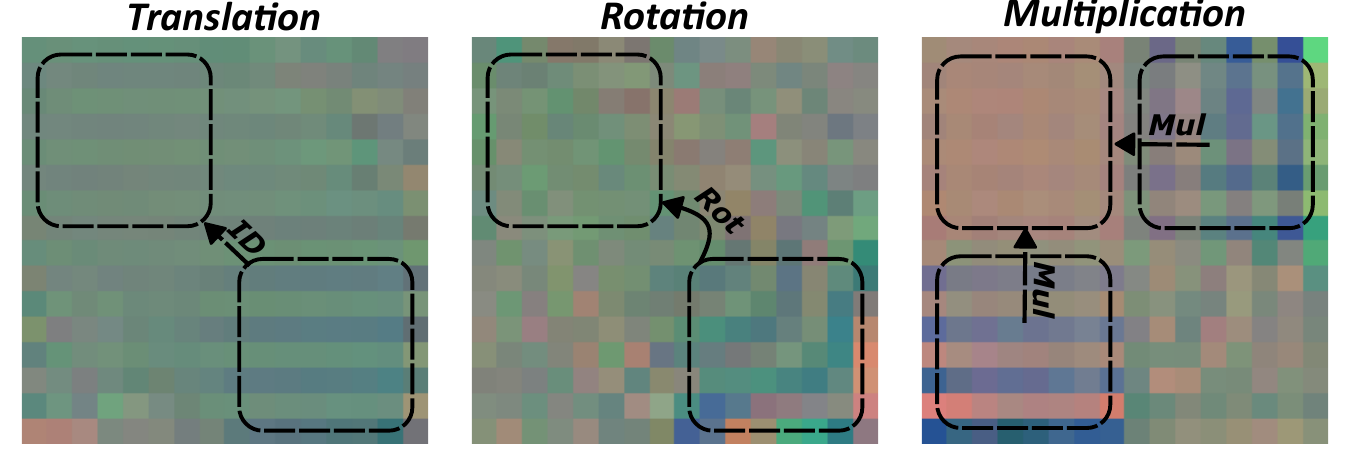}
    \caption{Monolithic hardware configurations for 3 different sub-tasks. We plot a PCA projection of the hidden channels to be able to display them as RGB. Colors thus does not have direct functional relevance. Overlayed are schematics of the (fixed) input-output transformations that each hardware was optimized on.}
    \label{fig:state_hw}
\end{figure}

Our first implementation optimizes task-specific parameters with the same spatial dimension as the computational state and a fixed number of hidden channels. This approach successfully trains the NCA on various tasks using specialized hardware for each. The optimized hardware configurations are visually interpretable, providing insights into the computation flow required for specific tasks (see \cref{fig:state_hw}).

However, this approach lacks generalizability. For example, hardware optimized for matrix translation from bottom-left to top-right would require retraining for the opposite operation. It also compromises the inherently scale-free nature of NCAs. An NCA trained for small-scale matrix multiplication would not be able generalize to larger matrices without hardware retuning. These limitations led us to develop a \textit{modular hardware approach}.

\subsubsection{Modular hardware}

To address the limitations of monolithic hardware, we developed a modular and composable approach. Similar to how specialized components on a motherboard emulate desired behavior, we train three purpose-specific hardware \textit{components}:

\begin{enumerate}
    \item An input embedding vector specifying cells that receive inputs in the computational task. Cells whose mutable state receives inputs include this vector in their immutable state.
    \item An output embedding vector marking cells that will serve as output during a specific task.
    \item A task embedding vector enabling the NCA to recognize the type of input-output transformation required. This learned vector is added to every cell's hardware state.
\end{enumerate}

These three core components are then manually assembled for each task instance to create specific task examples. The resulting immutable state remains accessible to the update rule throughout an experiment. This modular approach balances the scale-free nature of NCAs with the need for local heterogeneous substrate to perform diverse computational tasks. We also demonstrate that this enables zero-shot generalization, allowing the NCA to perform unseen task configurations and even composite task chaining (see \cref{subsec:future}).

\begin{figure}
    \centering
    \includegraphics[width=1.\linewidth]{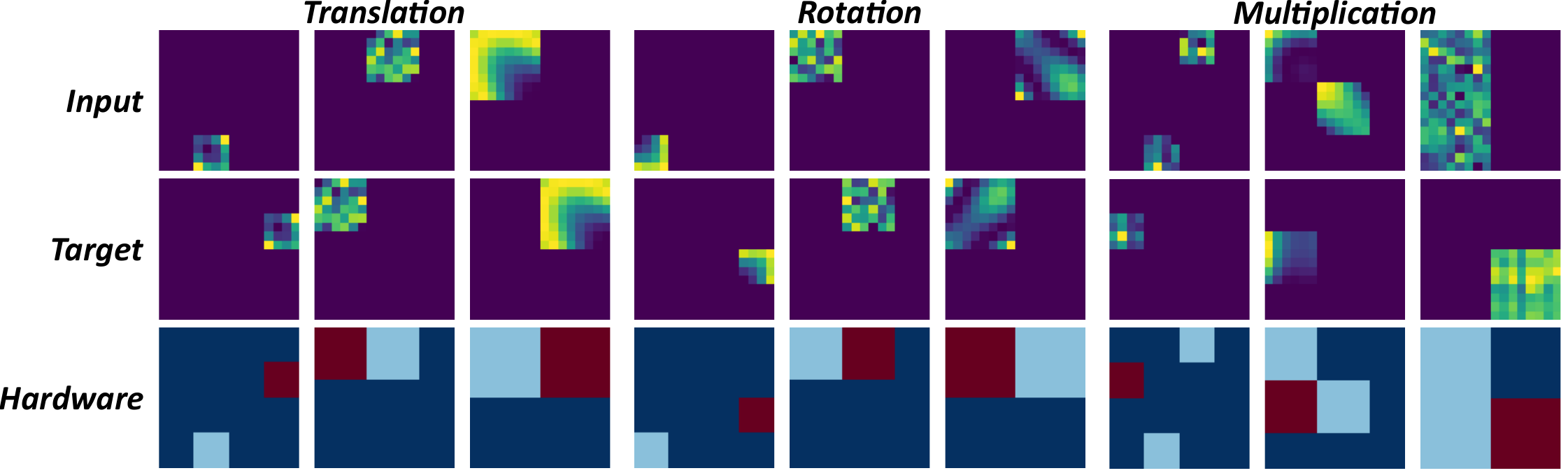}
    \caption{Modular hardware configurations for 3 different sub-tasks, and (now varying) different matrix sizes and placement.}
    \label{fig:modular_hw}
\end{figure}

\subsection{Tasks}
\label{subsec:tasks}

To train robust and versatile Neural Cellular Automata (NCA) models capable of performing general computation, we implement a flexible framework of matrix-based operations. These tasks, exercise different computational capabilities and test the NCA's ability to process, transform, and route information across the grid.

\paragraph{Matrix Operations}

The core of our framework is a set of fundamental matrix operations that represent different computational primitives:

\begin{itemize}
    \item \textbf{Identity Mapping}: Reproducing an input matrix at a different target location. This tests information preservation and signal routing across the grid.
    \item \textbf{Matrix Multiplication}: Given input matrices $A$ and $B$, compute $C = A \times B$. This tests the NCA's ability to perform non-local computations requiring information integration across regions.
    \item \textbf{Transposition /Rotation}: Given input matrix $A$, compute $B = A^T$ (or alternatively $B = A^T_{flipped}$ representing a 90 degree rotation in the plane) . This evaluates spatial information routing capabilities and geometric understanding.
\end{itemize}

\paragraph{Diverse Input Distributions}

To prevent overfitting to specific input patterns, we employ matrices with varied statistical properties:

\begin{itemize}
    \item \textbf{Uniform and Gaussian Distributions}: Randomly distributed matrix values test general processing capabilities.
    \item \textbf{Spatially Correlated Patterns}: Self-similar patterns with spatial correlations test the NCA's ability to process structured information and recognize spatial relationships.
    \item \textbf{Sparse Representations}: Matrices where most elements are close to zero test the NCA's efficiency in handling information sparsity.
\end{itemize}

\paragraph{Flexible Placement and Size}

To ensure the NCA develops robust computational abilities that generalize beyond fixed spatial arrangements, we include in the task framework:

\begin{itemize}
    \item \textbf{Dynamic Placement}: Input and output matrices can be positioned at different locations within the grid, preventing the NCA from memorizing fixed spatial patterns and forcing it to develop general computation mechanisms.
    \item \textbf{Variable Matrix Sizes}: Tasks involve matrices of different dimensions, from small to large relative to the grid size, testing the NCA's ability to scale its computations and adapt to varying information densities.
    \item \textbf{Multiple Inputs and Outputs}: Tasks can involve multiple input and output matrices distributed throughout the grid, requiring the NCA to coordinate information flow between different regions and perform parallel processing.
\end{itemize}

(This flexible task placement scheme is only usable in conjunction with a \textit{modular} hardware configuration, as discussed in \cref{subsec:hw})




Through this comprehensive framework, the NCA develops general computational capabilities that are robust to variations in task type, input distribution, matrix size, and spatial arrangement.

\subsection{Training}
\label{subsec:training}

To equip our Neural Cellular Automaton (NCA) with the capacity to perform a diverse range of computational tasks, we utilize a joint training framework. This approach simultaneously optimizes a single, shared NCA rule across a collection of distinct tasks instances. Each task is defined by a specific objective (an operation to be made on the inputs), and one task instance is typically represented by an initial grid state $S_0$, a desired target final state $S_{\text{target}}$, and often a mask $M$ indicating the regions of the grid relevant for evaluation.

During training, batches containing instances from various tasks are sampled. For each instance, the NCA model evolves the initial state $S_0$ over a defined number of discrete time steps $T_{\text{steps}}$ to produce a final state $S_{\text{final}}$. To enhance stability of the NCA, final states used to compute the loss are chosen at random between $T_{\text{steps}}$ and $T_{\text{steps}} - T_{\text{steps}} // 4$.

A loss function (commonly a masked error metric such as MSE), quantifies the discrepancy between the achieved final state and the target state within the relevant regions defined by the mask $M$.

Gradient-based optimization is employed to minimize this loss. Parameters associated with the shared NCA rule and any shared IO hardware components (in the \textit{modular} hardware case) are updated based on gradients aggregated across all tasks within the batch, promoting the learning of general-purpose computational primitives. Parameters belonging to task-specific modules (\textit{monolithic} hardware, or task components in the modular case) are updated using only the gradients derived from their corresponding task instances, enabling specialized behavior. This joint optimization process encourages the emergence of a versatile NCA capable of executing multiple computational functions through its learned local dynamics, dynamically adapting its behavior based on the presented task hardware.

\section{Experiments and Results}
\label{sec:results}

\subsection{Task Training}
\label{subsec:task_training}
\subsubsection{Joint Training}
In a multi-task training setup, our Neural Cellular Automata successfully master various matrix operations simultaneously through a shared update rule architecture combined with task-specific hardware components. Our findings demonstrate that a single NCA can develop general computational principles that apply across different matrix tasks while maintaining the specialized parameters needed for each specific operation.

The multi-task learning capability reveals the fundamental computational versatility of NCAs. By simultaneously learning to perform diverse operations such as matrix multiplication, translation, transposition and rotation within a unified framework, the model demonstrates mastery of a complete algebra of matrix operations—the essential building blocks for more complex computation.

This multi-task foundation directly enables more sophisticated composite applications, such as our MNIST classifier emulation (\cref{subsec:mnist}). The ability to decompose complex operations into smaller matrix tasks and process them through the same underlying cellular mechanism demonstrates a pathway toward increasingly complex computation. By establishing that NCAs can reliably perform these fundamental operations, we provide the essential building blocks for future work on more elaborate composite tasks, including full neural network emulation, algorithmic reasoning, and potentially even more advanced computational models implemented entirely within the cellular substrate.

\subsubsection{Downstream Tasks Fine-tuning}
A key advantage of our architecture emerges once the NCA is pre-trained: new tasks can be accommodated by fine-tuning only the hardware configurations while keeping the core CA update rules frozen. This approach significantly reduces computational requirements for adaptation to novel tasks. In our experiments, fine-tuning hardware alone increases training speed by a factor 2, compared to full model retraining. A more comprehensive comparison of joint training vs fine-tuning is nevertheless needed.

\subsection{MNIST Classifier emulation}\
\label{subsec:mnist}

\begin{figure*}
    \centering
    \includegraphics[width=0.8\linewidth]{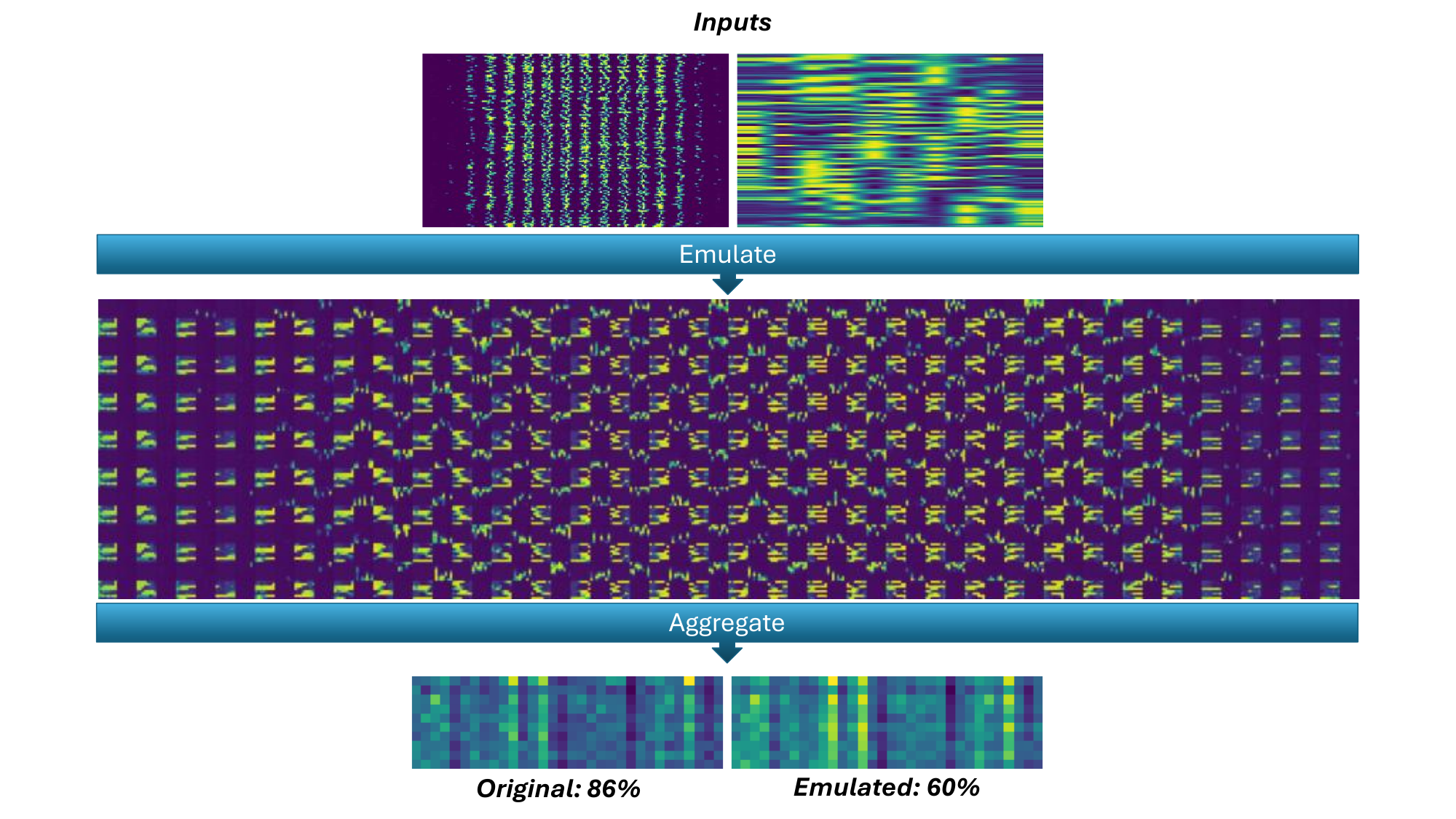}
    \caption{NCA emulates a neural network. Inputs show a batch of flattened MNIST images, alongside the weight matrix of a pre-trained single-layer linear classifier. We decompose this matrix multiplication into sub-blocks, that can be directly emulated in parallel by the NCA. Results are fetched from NCA states and aggregated back into logits (figure shows the first 32 outputs only for readability). We compute accuracy by taking the logits argmax per batch and comparing with labels.}
    \label{fig:mnist}
\end{figure*}

We demonstrate a practical downstream application by using our Neural Cellular Automata (NCA) to emulate an entire neural network directly in its computational workspace. Specifically, we emulate a single-layer Multi-Layer Perceptron (MLP) solving the MNIST digit classification task.

First, we pre-train a simple linear feedforward network to classify MNIST digits with sufficient accuracy. This classifier uses a single weight matrix without bias terms, where inference requires only a matrix multiplication between the flattened input images and the weight matrix, followed by an argmax operation to determine the predicted digit.
Our NCA model was pre-trained as well, on smaller 8×8 matrix multiplication tasks. While we could hope for generalization to larger matrices, operations of the scale required for MNIST classification (784×10) would exceed the capacity of what can be performed by such a model. To address this limitation, we implement block-matrix decomposition, fragmenting the classification of MNIST images into multiple smaller 8×8 matrix operations that fit within the NCA's state constraints.
The resulting decomposed operations can be executed directly within the NCA's computational state without requiring task-specific fine-tuning, demonstrating the robustness of our approach to novel matrix distributions. The NCA processes each block multiplication operation in parallel, after which we aggregate the results to reconstruct the complete classification logits.
When evaluating performance, we compare the predictions and accuracy of our NCA-based emulation against the original classifier. While we observe some accuracy degradation due to error propagation across numerous sub-operations, the model still achieves respectable performance (around $60 \%$ accuracy for the emulated classification compared to $84 \%$ for the original, with predictions agreeing around $69 \%$ of the time). We argue that this is providing empirical evidence that neural network emulation via NCA is feasible.

This research has significant implications for analog and physical computing. If our NCA's update rules were implemented as physical state transitions, this would represent a pathway toward physical neural network emulation without reverting to binary-level operations. The ability to operate directly at the level of matrix operations using hardware specifically designed for this computational paradigm could offer substantial efficiency and performance improvements over conventional digital approaches.

\subsection{Future directions: task composition and neural compiler}
\label{subsec:future}



\begin{figure}
    \centering
    \includegraphics[width=0.9\linewidth]{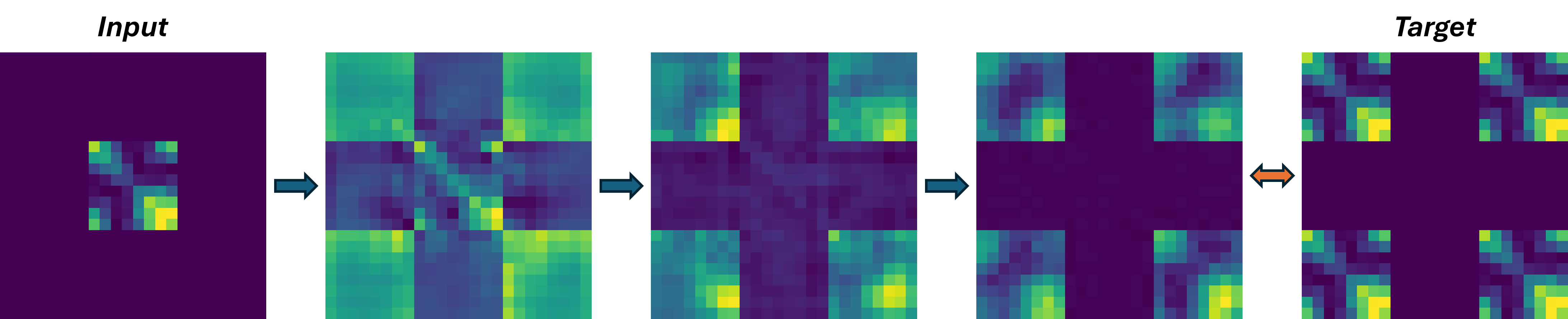}
    \caption{Examples of an Out of Distribution Task: the NCA needs to distribute a matrix in all corners in a larger grid than the one seen during training.}
    \label{fig:OOD}
    \Description{}
\end{figure}
\begin{figure*}
    \centering
    \includegraphics[width=0.99\linewidth]{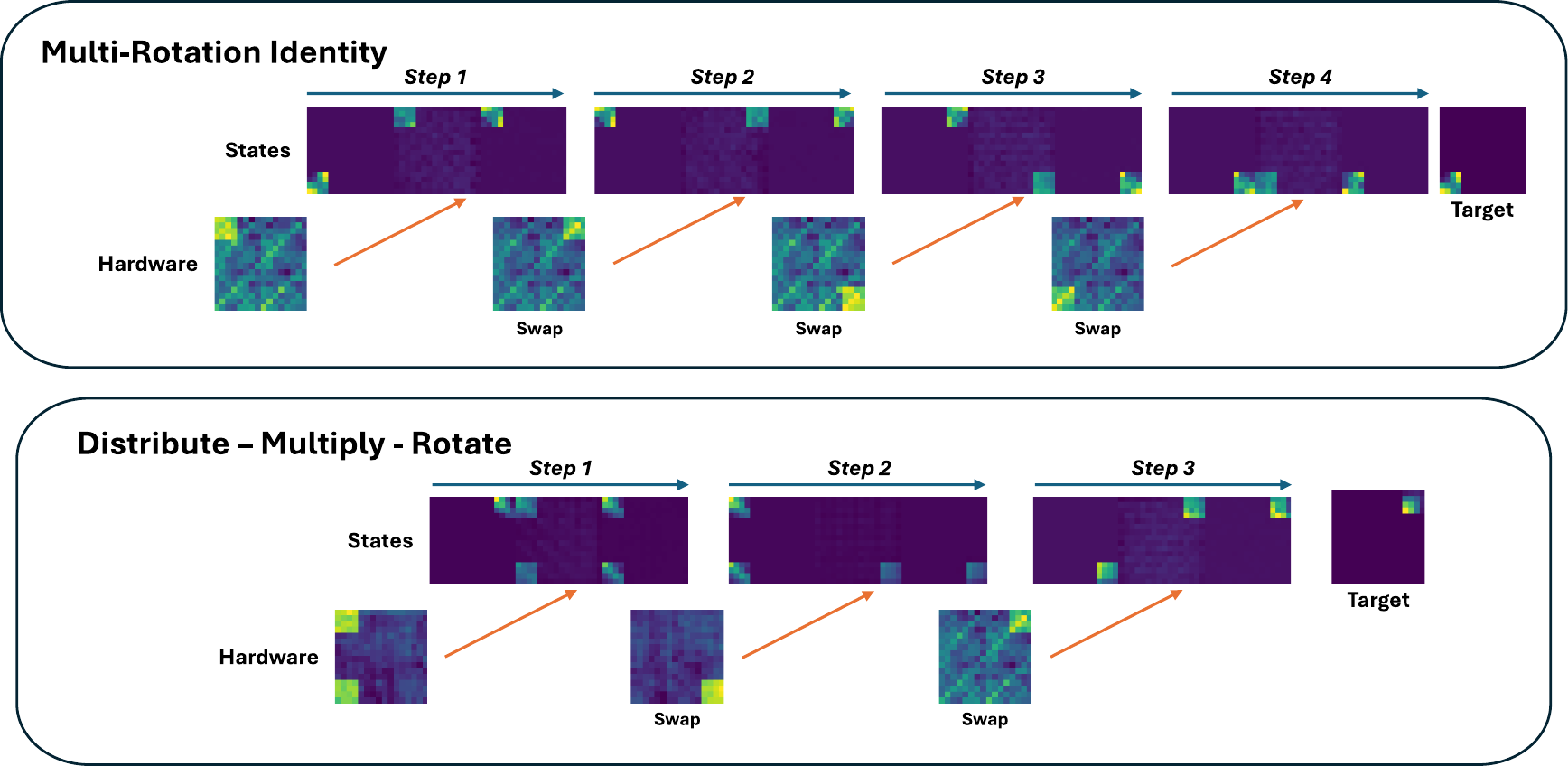}
    \caption{Illustrating composite computational tasks using modular hardware configurations. Top panel: Multi-Rotation Sequential rotations resolving in the identity function. Bottom panel: Distribute-Multiply-Rotate task, showing a three-step process where matrices are distributed, multiplied, and rotated to achieve the target state. Each step shows both the hardware configuration and corresponding computational states. }
    \label{fig:compiler}
    \Description{}
\end{figure*}

The modular hardware configuration we developed enables straightforward creation of out-of-distribution tasks through component composition. This flexibility allows us to design novel computational scenarios that the NCA was not explicitly trained on, yet can still execute successfully.
For instance, we can implement data distribution patterns by duplicating a central matrix into multiple corners using the matrix translation task embedding and multiple target tiles \cref{fig:OOD}. This simple example demonstrates how our architecture supports operations beyond the training distribution without requiring additional training.
This framework opens the path toward complex composite tasks created through sequential chaining of primitive operations. Consider the following multi-step procedure (\cref{fig:compiler} bottom panel):
\begin{enumerate}
    \item Start with a input matrix and distribute copies to two corner positions using target tiles.
    \item Replace the current hardware configuration with new parameters that redefine these targets as inputs, then perform matrix multiplication towards a third corner.
    \item Update the hardware again to rotate the resulting matrix and return it to the original position.

\end{enumerate}

While such composite sequences may not appear useful on their own, they demonstrate a critical capability: the NCA can execute complex algorithmic workflows through sequential reconfiguration of its hardware parameters. This capability lays the groundwork for more sophisticated computational reasoning and abstraction.

This composite task chaining also highlights the crucial role of stability in achieving composable computations. When outputs produced within the NCA's computational state serve as inputs for subsequent operations, establishing stable representations becomes essential. In parallel to biological systems, where homeostasis maintains internal equilibrium despite external changes, NCAs require a form of computational homeostasis to maintain reliable state representations between operations. Similarly, analog computers may require homeostasis to successfully implement extended chains of tasks and computations without degradation of information.

We propose that this sequential operation model suggests a compelling dual-timestep approach to neural compilation:
At the neuronal timestep, the NCA's mutable state evolves according to its update rules, creating the fundamental dynamics of computation.
At the compiler timestep, hardware parameters are reconfigured to provide task abstractions and high-level procedural steps.
This separation of concerns, where fast neuronal dynamics handle computation while slower hardware changes control program flow, mirrors classical computer architecture but within a continuous, differentiable substrate. As this approach matures, it could enable direct compilation of algorithms into neural cellular automata, combining the flexibility of neural networks with programmatic execution. This would possibly be facilitated by a better task-abstraction and hardware generation that we detail in the next section.

\subsubsection{Graph-based hardware hypernetwork}
Finally, building upon the limitations of the previous hardware approaches, we are curently developing a more principled graph-based hardware generation framework that offers significant improvements in both flexibility and scale-invariance. This (WIP) approach leverages a task representation abstraction where computational operations are modelled as a graph, with nodes representing input and output regions and edges encoding specific transformations between them.
At the core of this framework is a Hardware Meta-Network consisting of two main components: a Graph Neural Network (GNN) encoder and a coordinate-based hypernetwork.

The GNN processes a task graph structure, where nodes contain normalized spatial information about input and output regions, and edges represent specific operations (e.g., matrix multiplication, rotation) to be performed between regions. Through multiple message-passing layers, the GNN distils this graph representation into a fixed-dimensional latent task vector that captures the essential computational requirements of a single task instance problem.

This latent representation then conditions a coordinate MLP hypernetwork that generates hardware vectors for every spatial location in a scale-free manner. The hypernetwork leverages positional encodings to create spatially varying hardware patterns that guide the NCA's computational dynamics across the grid. Crucially, this approach maintains exact spatial invariance: task specifications are normalized relative to grid dimensions, which should enable the generated hardware to automatically adapt to different grid sizes and region placements without retraining.

\begin{figure}[h!]
    \centering
    \includegraphics[width=0.99\linewidth]{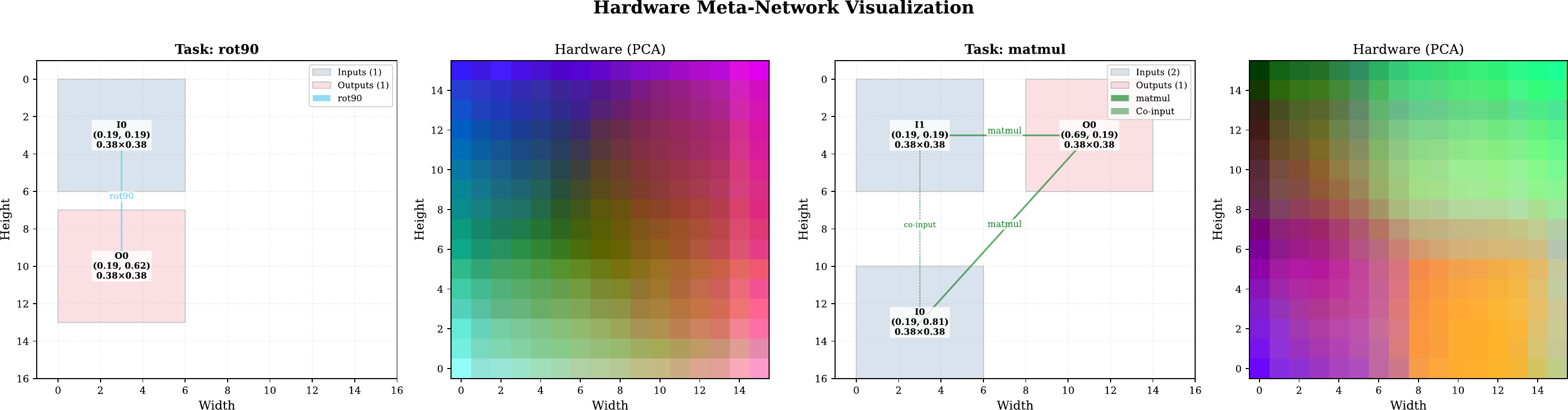}
    \caption{Graph-based tasks representations and GNN-based hypernetwork for hardware generation }
    \label{fig:gnn}
\end{figure}

This graph-based representation provides an intuitive interface between human-specified computational tasks and the continuous NCA substrate. Users can define tasks through a natural graph specification (inputs, outputs, and operations), and the meta-network translates these specifications into appropriate hardware configurations. This approach would effetively improve on the concept of a compiler between human intent and the NCA's computational capabilities, allowing for better definition of task chaining and temporal hardware evolution. Moreover, the graph structure can enable rich extensions beyond our initial implementation. Tasks can be ordered through edge attributes, allowing sequential execution planning. Dynamic hardware reconfiguration becomes possible by modifying the task graph over time, creating a secondary dynamics layer that complements the fast neural dynamics of the cellular automaton itself. This hierarchical temporal structure, where fast neural dynamics implement local computations while slower hardware dynamics guide algorithmic flow, mirrors the dichotomy in traditional computing architectures between clock-cycle operations and higher-level program execution. Doing so within a unified differentiable framework may ultimately enable more efficient, adaptable continuous computational paradigms.

\section{Conclusion}
The exploration of universal computation within cellular automata has historically been confined to discrete systems, where models like Conway’s Game of Life and Elementary Cellular Automata have demonstrated the remarkable ability to emulate Universal Turing Machines. However, extending this capability to continuous cellular automata presents significant challenges, primarily due to the absence of discrete states and the inherent instability of smooth, analog dynamics. In this work, we have taken pragmatic first steps toward overcoming these hurdles by leveraging NCA as a substrate for developing universal computation in a continuous domain. By employing gradient descent to train NCA rules, we have demonstrated a pathway to sculpt complex computational behaviours without the need for manual rule design, shifting the burden of discovery from human ingenuity to machine learning.

Our results illustrate that NCA can successfully encode fundamental computational primitives, such as matrix multiplication and inversion, and even emulate a neural network capable of solving the MNIST digit classification task directly within its state. These findings suggest that NCAs can serve as a bridge between traditional computing architectures and self-organizing systems, offering a novel computational paradigm that aligns closely with analog systems. This linkage is particularly promising for designing efficient computational frameworks for AI models operating, where energy efficiency and robustness are paramount. Rather than training entirely new rules for each task, our approach hints at the possibility of discovering optimal hardware configurations that exploit the fixed physical laws governing these substrates, enabling meaningful computations with minimal overhead.

Looking forward, we believe this work lays the groundwork for transformative advancements in computational science. By automating the discovery of general-purpose computers within diverse physical implementations, NCA could revolutionize how we harness novel materials and systems for computation, potentially leading to ultra-efficient analog hardware systems or computational paradigms that scale linearly with resource demands. While challenges remain --- such as stabilizing continuous dynamics for reliable symbolic encoding and scaling these systems to more complex tasks—the potential of NCA to unlock universal computation in continuous cellular automata opens new avenues for exploration. Ultimately, this research not only advances our understanding of computation in continuous dynamics but also paves the way for the next generation of adaptive, energy-efficient computing technologies.

\bibliographystyle{ACM-Reference-Format}
\bibliography{ref}



\end{document}